\documentclass[10pt, conference, compsocconf]{IEEEtran} 

\usepackage{amsmath} 
\usepackage{subfigure}
\usepackage{amsmath} 
\usepackage[pdftex]{graphicx}
\usepackage{subfigure}
\usepackage{float}
\usepackage{color}
\usepackage[utf8]{inputenc}    
\usepackage[T1]{fontenc}       
\usepackage{hyperref}

\usepackage[backend=biber, style=ieee, sorting=none]{biblatex}
\AtBeginBibliography{\small}
\begin{document}

\title{\vspace*{-0.4in} \LARGE \bf aUToTrack: A Lightweight Object Detection and Tracking System for the SAE AutoDrive Challenge \vspace*{-0.4in} }
\author{\IEEEauthorblockN{Keenan Burnett, Sepehr Samavi, Steven L. Waslander, Timothy D. Barfoot, Angela P. Schoellig}
\IEEEauthorblockA{Institute for Aerospace Studies\\
University of Toronto\\
Toronto, Canada\\
\{keenan.burnett, sepehr\}@robotics.utias.utoronto.ca}
}


\maketitle
\vspace*{-0.4in}
\begin{abstract}
    The University of Toronto is one of eight teams competing in the SAE AutoDrive Challenge -- a competition to develop a self-driving car by 2020. After placing first at the Year 1 challenge \cite{burnett2018building}, we are headed to MCity in June 2019 for the second challenge. There, we will interact with pedestrians, cyclists, and cars. For safe operation, it is critical to have an accurate estimate of the position of all objects surrounding the vehicle. The contributions of this work are twofold: First, we present a new object detection and tracking dataset (UofTPed50), which uses GPS to ground truth the position and velocity of a pedestrian. To our knowledge, a dataset of this type for pedestrians has not been shown in the literature before. Second, we present a lightweight object detection and tracking system (aUToTrack) that uses vision, LIDAR, and GPS/IMU positioning to achieve state-of-the-art performance on the KITTI Object Tracking benchmark. We show that aUToTrack accurately estimates the position and velocity of pedestrians, in real-time, using CPUs only. aUToTrack has been tested in closed-loop experiments on a real self-driving car (seen in Figure~\ref{fig:sensors_diagram}), and we demonstrate its performance on our dataset.
\end{abstract}

\section{Introduction}

Standard object detection and tracking algorithms used for video understanding use 2D bounding boxes to identify objects of interest. For autonomous driving, 2D bounding boxes are insufficient. A 3D position and velocity estimate is required in order to localize objects in a map and anticipate their motion. Existing object detection benchmarks compare detector outputs against hand-generated labels either in 2D image coordinates or in a 3D sensor frame \cite{KITTI}. In both cases the "ground truth" has been generated by a human with some semi-automatic tools. For this reason, such datasets are subject to human biases in the labelling process.

Current leading methods on these benchmarks do so by replicating the labels observed in the training set \cite{ren2017accurate}. Unfortunately, this does not necessarily mean that these approaches are accurately localizing objects in 3D space, which is the critical problem that we address here.

Existing datasets \cite{KITTI} lack a means for benchmarking 3D Object Detection and Tracking for pedestrians. To address this shortcoming, we introduce UofTPed50, a new dataset that we are making publicly available in June 2019. UofTPed50 includes vision, LIDAR, and GPS/IMU data collected on our self-driving car in 50 sequences involving interactions with a pedestrian. We use a separate GPS system attached to the pedestrian to obtain ground truth positioning. By using GPS ground truth instead of hand labels, we can rigorously assess the the localization accuracy of our system. To our knowledge, benchmarking pedestrian localization in this manner has not been shown in the literature before.


\begin{figure} [t]
    \includegraphics[width=\columnwidth]{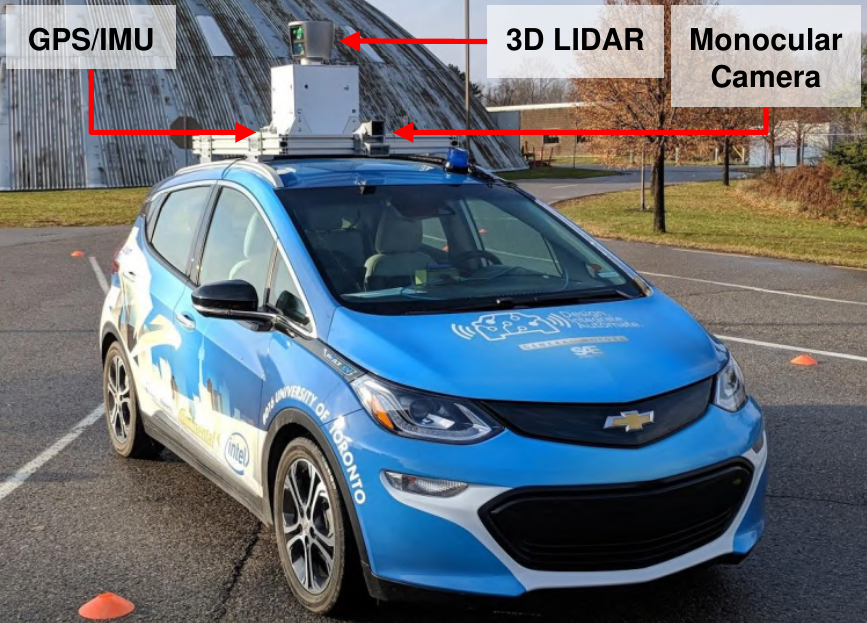}    
    \centering
    \caption{Our self-driving car \textit{Zeus} at the University of Toronto Institute for Aerospace Studies (UTIAS). \url{https://youtu.be/FLCgcgzNo80}}
    \label{fig:sensors_diagram}
\end{figure}

\begin{figure*} [ht]
    \includegraphics[width=\linewidth]{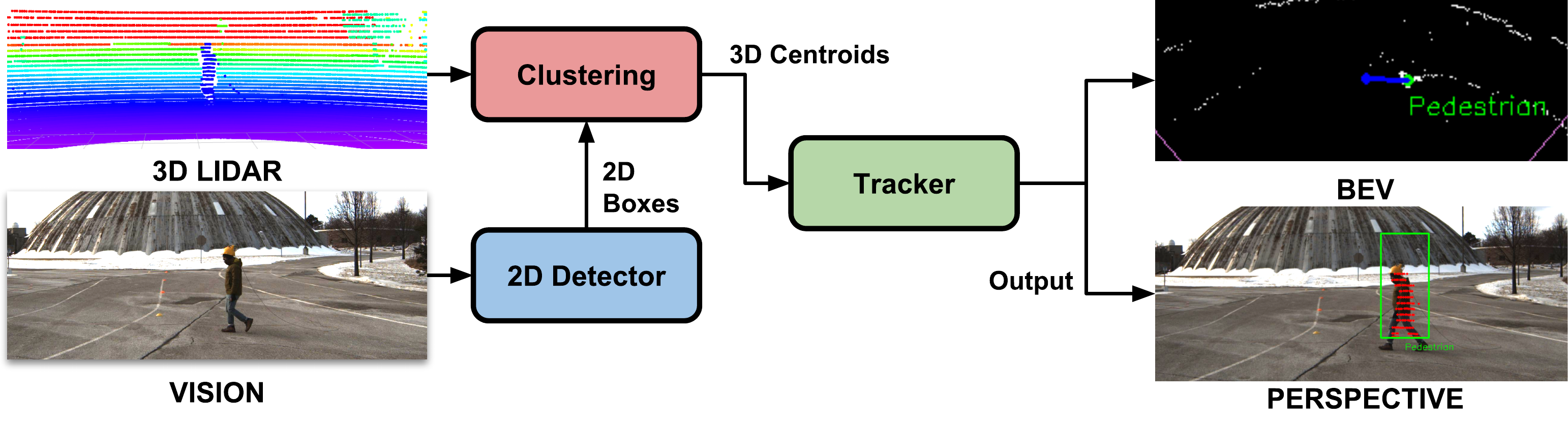}
    \centering
    \caption{Our pipeline for 3D object detection and tracking. For UofTPed50 experiments, SqueezeDet is our 2D Detector. Clustering consists of Euclidean clustering over the LIDAR points that fall within the bounding boxes when projected onto the image plane. Our tracker consists of gated nearest neighbor data association and an Extended Kalman Filter for tracking position and velocity in 3D space.}
    \label{fig:ros_diagram}
\end{figure*}

As a secondary contribution, we describe our approach to Object Detection and Tracking (aUToTrack), and demonstrate its performance on UofTPed50 and KITTI. aUToTrack consists of an off-the-shelf vision-based 2D object detector paired with a LIDAR clustering algorithm to extract a depth for each object. GPS/IMU data is then used to localize objects in a metric reference frame. Given these 3D measurements, we use greedy data association and an Extended Kalman Filter (EKF) to track the position and velocity of each object. Figure~\ref{fig:ros_diagram} illustrates the aUToTrack pipeline. We demonstrate state-of-the-art performance on the KITTI Object Tracking benchmark while using data association and filtering techniques that are faster and much simpler than many competing approaches \cite{choi2015nomt}, \cite{sharma2018beyondpixels}, \cite{xiang2015learning}, \cite{scheidegger2018mono}. We also demonstrate that we can accurately estimate the position and velocity of pedestrians using our UofTPed50 dataset while running our entire pipeline in less than 75 ms on CPUs only.

Estimating the velocity of objects is a difficult problem in self-driving \cite{cruise} which we tackle in this work. Systems like ours that are lightweight, and capable of running on CPUs are uncommon in the literature but essential in practice. Although aUToTrack was designed for the SAE AutoDrive Challenge, it has utility for many robotic systems deployed in human-centered domains.

\section{Related Work}
\subsection{Object Detection}


Object detectors can be classified into 2D and 3D detectors.
Among 2D detectors, two-stage detectors have historically achieved the top accuracy on public benchmarks \cite{fasterrcnn}. Single-stage detectors such as YOLO \cite{liu2016ssd} and SSD \cite{redmon2016you} tend to be more computationally efficient, but typically do not achieve the best performance. A recent work that focuses on achieving a high framerate on limited hardware is SqueezeDet \cite{wu2017squeezedet}. A work that focused on achieving the best possible accuracy is Recurrent Rolling Convolutions (RRC).

3D detectors estimate the centroid and volume of objects using a 3D bounding box. Recent works in this area include {Frustum PointNets} \cite{qi2017frustum} and {AVOD} \cite{avod}. Frustum PointNets uses a 2D detector to cut out a "frustum" from an incoming pointcloud. They then use a specialized fully-connected network to cluster points within the frustum and regress a 3D bounding box. {AVOD} fuses vision- and LIDAR-based features together in both the region proposal and a bounding box regression stage of their network.

For the purposes of self-driving, the question becomes: is it better to use a 2D or a 3D detector? 2D object detection is perceived to be a more mature field, with many approaches exceeding 90\% mean Average Precision (mAP) at acceptable framerates \cite{KITTI}. On the other hand, 3D detections are more useful for self-driving since they incorporate an accurate centroid estimate which can be used directly to safely interact with traffic participants. However, most of these detectors are computationally very expensive. AVOD, which is one of the faster 3D detectors, requires a TITAN X GPU to achieve an inference time of 100 ms. For this reason, the decision to use a fast single-stage detector, in our case SqueezeDet, becomes clear. Compared with other approaches in literature, we are able to estimate accurate 3D centroids for pedestrians using CPUs only. This is uncommon in literature and enables our system to be used in a real self-driving car.

\subsection{Object Tracking}

Recent works in the area of 2D object tracking include \cite{choi2015nomt}, \cite{sharma2018beyondpixels}, \cite{sadeghian2017learningtrack}, \cite{bewley2016simple}. In \cite{choi2015nomt}, an aggregated local flow descriptor is used to associate targets and detections over a temporal window. In \cite{sharma2018beyondpixels}, geometry, object shape, and pose costs are used to augment data association. In \cite{sadeghian2017learningtrack}, the authors formulate multi-object tracking using Markov Decision Processes.

The above approaches score high on KITTI, however each of their runtimes exceeds 0.2 s, preventing their use in a self-driving car. Our approach is simpler and we are able to achieve competitive performance with lower runtime.

In \cite{bewley2016simple}, the authors describe a fast 2D object tracker that consists of a Kalman Filter and the Hungarian algorithm. For 2D tracking, they achieve competitive results on the MOT benchmark \cite{leal2015motchallenge}, while being much faster than competitors.

In 3D object detection and tracking, recent works include \cite{moosmann2013joint}, \cite{dewan2016motion}. In \cite{moosmann2013joint}, a model-free object detection scheme based on convexity segmentation is used with a Kalman filter and constant velocity motion model. In \cite{dewan2016motion}, a model-free 3D object detection and tracking method based on motion cues is used. Both approaches use the same dataset, which has a sensor vehicle equipped with a LIDAR and a GPS/IMU as well as a separate GPS sensor used to ground truth the target vehicle. These approaches both have several drawbacks. First, LIDAR-based approaches are not able to classify different objects easily. In addition, \cite{dewan2016motion} struggles to detect pedestrians, partly because of their motion-based detections. Neither approach presents results for pedestrian detection. The dataset presented in this work has more variation in target distance, velocities, and trajectories and is organized for a rigorous evaluation of an object detection and tracking system.

A recent work similar to ours is \cite{scheidegger2018mono}. In this work, the authors employ a DNN trained to detect objects and estimate depth from a single image. For tracking, they use a Poisson Multi-Bernoulli Mixture filter. They achieve competitive results on KITTI using only images. However, when LIDAR is readily available, approaches such as ours perform better.

\section{2D Object Detector}


The SAE AutoDrive Challenge does not allow GPUs for on-board computations. Thus, we have invested significant effort into selecting a lightweight object detector capable of running on CPUs. With this in mind, we have chosen to use squeezeDet \cite{wu2017squeezedet} as our 2D detector for real-time experiments on UofTPed50. SqueezeDet achieves top-25 accuracy for pedestrians, and top-75 accuracy for cars on KITTI. In addition, it is one of the fastest approaches, achieving a framerate of 57 FPS on a TITAN X GPU (1248x384 images). SqueezeDet is also fully-convolutional, which allows for run-time optimization using Intel's OpenVINO Acceleration Tools \cite{openvino}. We are able to run SqueezeDet at 31 FPS using only 8 CPU cores (1248x384 images). 

For experiments on the KITTI Tracking benchmark, we use Recurrent Rolling Convolutions (RRC) \cite{ren2017accurate} since it is ranked top among published works on the 2D Object Detection benchmark. RRC is substantially slower than SqueezeDet, requiring over 1500 ms to process a single frame on a GTX1080Ti. This prevents its use in real-time experiments. However, the increase in recall on KITTI is substantial: from 65\% with SqueezeDet to 94\% with RRC. 
%
%
%
\section{Clustering}
In order to achieve real-time performance, we restrict our attention to points in front of the vehicle up to 40m away, and 15m to each side. We subsequently segment and extract the ground plane. The remaining points are transformed into the camera frame. A corresponding set of image plane locations are obtained by projecting LIDAR points onto the image plane using an ideal perspective projection model and the intrinsic camera matrix.
%
%

%
%
%
We then retrieve the LIDAR points that lie inside each 2D bounding box. We then use Euclidean clustering on the corresponding 3D LIDAR points \cite{Rusu_ICRA2011_PCL}. Several heuristics are used to choose the best cluster from the clustering process. These heuristics include: comparing the detected distance to the expected size of the object and counting the number of points per cluster. Once the best cluster has been chosen, the position of the object is computed as the centroid of the points in the cluster. 

In section~\ref{sec:experiments}, we show that this method of centroid estimation works well for pedestrians. However, this method is not as accurate for cars. This is likely because the majority of the points being clustered are from the face of the object facing the LIDAR. For this reason, our centroid estimation may be off by 1 m or more. In our qualitative analysis of results on public road driving, we have observed that our approach is still effective at estimating the position and velocity of parked cars and vehicles travelling longitudinally with respect to the sensor vehicle. These estimates may be used to enable autonomous highway driving. However, for safe autonomous driving on a a crowded roadway, a more accurate centroid estimate is likely required.

\section{Tracker Setup}
\label{sec:trackerSetup}
For each object, we keep a record of the state, $\hat{\textbf{x}}$, covariance, $\hat{\textbf{P}}$, \textit{class, confidence level}, and \textit{counters} for track management. The state is defined in Equation~\eqref{obj-state}, where $(x,y,z)$ is the position, $(\dot{x},\dot{y})$ is the velocity within the ground plane, and $(w,h)$ are the bounding box width and height in the image plane. The position and velocity are tracked in a static map frame external to the vehicle. The class, which is output by SqueezeDet can be one of: $(car, pedestrian, cyclist)$. For track management, we count the number of frames an objects has gone without being observed, and a boolean for "trial" objects.

\subsection{Constant Velocity Motion and Measurement Models}
The following equations describe the constant velocity motion model that is used for all detected objects:
\begin{align}
    \textbf{x} &= \begin{bmatrix} x & y & z & \dot{x} & \dot{y} & w & h \end{bmatrix}^T \label{obj-state}\\
    \textbf{y} &= \begin{bmatrix} x & y & z & w & h \end{bmatrix}^T \\
    \textbf{x}_k &= \textbf{A}\textbf{x}_{k-1} + \boldsymbol{\omega} \\
    \textbf{y}_k &= \textbf{C}\textbf{x}_{k} + \boldsymbol{n} \\
    \boldsymbol{\omega} &\sim \mathcal{N}(\textbf{0}, \textbf{Q}) \\
    \boldsymbol{n} &\sim \mathcal{N}(\textbf{0}, \textbf{R})\\
    \textbf{A} &= 
        \begin{bmatrix}
            \textbf{1}_2 & \textbf{0}_{2\times1} & T*\textbf{1}_2 & \textbf{0}_{2}\\
            \textbf{0}_{5\times2} & \textbf{1}_5
        \end{bmatrix} \\
    \textbf{C} &= 
        \begin{bmatrix}
            \textbf{1}_3 & \textbf{0}_{3\times4} \\
            \textbf{0}_{2\times5} & \textbf{1}_2
        \end{bmatrix}
\end{align}
In this setup, \textbf{x} is the state of the object, \textbf{y} is the measurement, $\boldsymbol{\omega}$ is the system noise, $\boldsymbol{n}$ is the measurement noise, \textbf{A} is the state matrix, and \textbf{C} is the observation matrix. \textbf{Q} and \textbf{R} are the system noise and measurement noise covariances respectively, which we assume to be diagonal. The diagonal entries of these matrices are used to tune the tracker for responsiveness vs. smoothness.

\subsection{Linear Kalman Filter}
The following equations describe the linear Kalman filter used to track the state and covariance of each object: 
\begin{align}
    \check{\textbf{x}}_k &= \textbf{A} \hat{\textbf{x}}_{k-1} \label{pred1}\\
    \check{\textbf{P}}_k &= \textbf{A} \hat{\textbf{P}}_{k-1} \textbf{A}^T + \textbf{Q}  \label{pred2}\\
    \textbf{K}_k &= \check{\textbf{P}}_{k} \textbf{C}^T (\textbf{C} \check{\textbf{P}}_{k}  \textbf{C}^T + \textbf{R})^{-1} \label{kgain}\\
    \hat{\textbf{P}}_k &= (\textbf{1} - \textbf{K}_k \textbf{C}) \check{\textbf{P}}_k \label{correct1}\\
    \hat{\textbf{x}}_k &= \check{\textbf{x}}_k + \textbf{K}_k ( \textbf{y}_k - \textbf{C} \check{\textbf{x}}_k) \label{correct2}
\end{align}
In this setup, Equations (\ref{pred1})-(\ref{pred2}) are the prediction, Equation \eqref{kgain} is the Kalman gain, and Equations (\ref{correct1})-(\ref{correct2}) are the corrector equations.

\subsection{Tracking in World Frame}
\label{sec:trackingWorld}
We require the position and velocity of objects in a static map frame denoted $o$. Raw measurements are given in a sensor frame attached to the vehicle, denoted $c2$. We augment some of our vectors in order to track in the static frame as follows:






\begin{align}
    \textbf{x}' &= \begin{bmatrix} x & y & z & 1 & \dot{x} & \dot{y} & w & h \end{bmatrix}^T \\
    \textbf{C}' &= 
        \begin{bmatrix}
            \textbf{1}_3 & \textbf{0}_{3\times3} & \textbf{0}_{3\times2} \\
            \textbf{0}_{2\times3} & \textbf{0}_{2\times3} & \textbf{1}_2
        \end{bmatrix} \\
    \textbf{T}_{c_2 o}' &= 
        \begin{bmatrix}
            \textbf{T}_{c_2 o} & \textbf{0}_{4\times4} \\
            \textbf{0}_{4\times4} & \textbf{1}_4
        \end{bmatrix} \\
    \textbf{G}_k &= \textbf{C}' \textbf{T}_{c_2 o}' \\
    \textbf{y} &= \textbf{C}' \textbf{T}_{c_2 o}' (\textbf{x}_k' + \boldsymbol{n}')
\end{align}

Keeping our prediction equations the same, our update equations become:

\begin{align}
\textbf{K}_k &= \check{\textbf{P}}_{k}' \textbf{G}_k^T (\textbf{G}_k \check{\textbf{P}}_{k}' \textbf{G}_k^T + \textbf{R})^{-1} \\
\hat{\textbf{P}}_k' &= (\textbf{1} - \textbf{K}_k \textbf{G}_k) \check{\textbf{P}}_k' \\
\hat{\textbf{x}}_k' &= \check{\textbf{x}}_k' + \textbf{K}_k ( \textbf{y}_k - \textbf{G}_k \check{\textbf{x}}_k')
\end{align}




We then remove augmented components of $\hat{\textbf{P}}_k'$, $\hat{\textbf{x}}_k'$ at the end before publishing our answer to other components of the system.

\section{Data Association}

Our data association is based on metric positions only and does not use the appearance of objects. We have found this to work well for 3D object measurements since objects such as cars tend to be separated by significant distances. We use static gates to associate new detections to existing tracks. The gates are constructed using the maximum possible inter-frame motion. For example, assuming a maximum speed of 6 m/s for pedestrians and a maximum speed of 20 m/s for vehicles and a 0.1 s time step, we have a gating region of 0.6 m and 2.0 m respectively. To account for the high speeds of cars, we use the tracked velocity of each car to update its gate radius.
We use a greedy approach to associate measurements to tracked objects. For each tracked object, we evaluate the object's distance to the observations that are within its gate and find the nearest neighbour to the object among the gated observations. We associate the measurement with the tracked object and remove the measurement from the list. We repeat the process for every object that has detections in its gate.

\begin{figure*} [t!]
    \includegraphics[width=\linewidth]{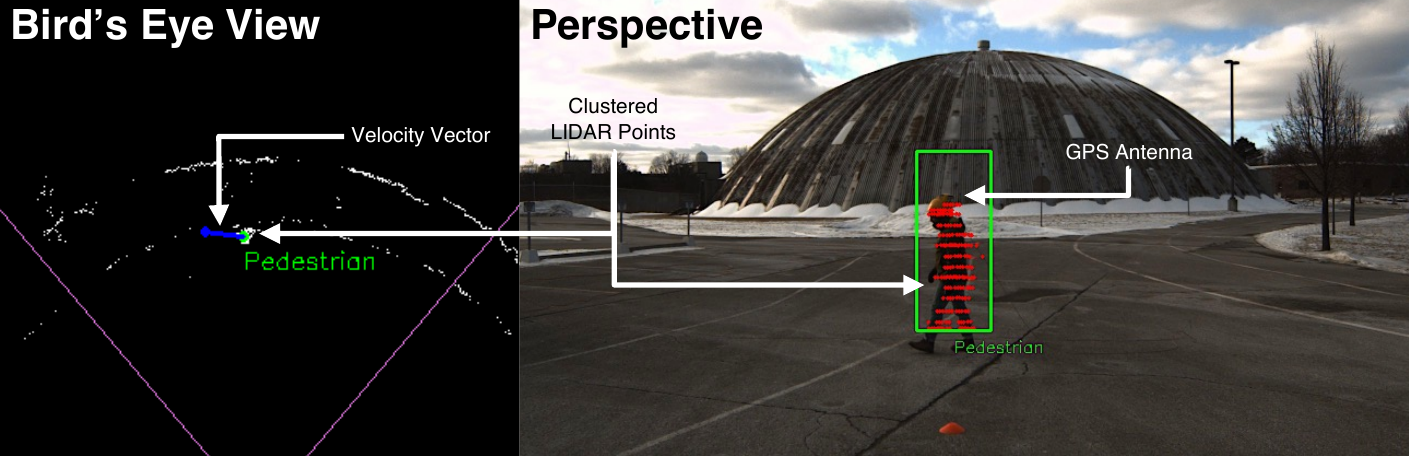} 
    \centering
    \caption{Example of a pedestrian state estimate on the UofTPed50 dataset. On the left, we plot a bird's eye view with LIDAR points projected to the ground. On the right, we plot a perspective view from the camera.}
    \label{fig:rqt_output}
    \vspace{-5mm}
\end{figure*}
\section{Managing The List of Tracked Objects}
We employ a strategy of greedy track creation and lazy deletion for managing tracks. In greedy track creation, every observation becomes a new track. However, every track must go through a "trial period".  While objects are in their trial period, they are removed from the list of objects being tracked if they miss a single frame.

Once objects are promoted from their trial period, we count the number of consecutive frames that an object has been unobserved for. In order for a non-trial track to be removed from the list, there must be no associated measurements for Y (Y=5) consecutive frames. This constitutes a lazy deletion process.

Table \ref{table:runtime} summarizes our entire pipeline's run-time. We note that a significant portion of the run-time can be attributed to the 2D detector.  

\begin{table}[t]
    \centering
    \caption{Runtime of Each Component in our Pipeline.}
    \label{runtime}
    \begin{tabular}{|l||l|l|}
    \hline
    \textbf{Component}       & \textbf{Run Time} & \textbf{Hardware}              \\ \hline \hline
    SqueezeDet & 18 ms              & GTX1080Ti                \\ \hline
    SqueezeDet & 26 ms              & Arria 10 FPGA                \\ \hline
    SqueezeDet & \textbf{32 ms}             & Intel Xeon E5-2699R (8 cores) \\ \hline
    Clustering               & \textbf{30 ms}             & Intel Xeon E5-2699R (1 core)   \\ \hline
    Tracker                  & \textbf{10 ms}              & Intel Xeon E5-2699R (1 core)   \\ \hline
    \textbf{Total Run Time:} & \textbf{72 ms}             & Intel Xeon E5-2699R Only       \\ \hline
    \end{tabular}
    \label{table:runtime}
\end{table}


\section{UofTPed50}


As the primary contribution of this work, we are releasing a new dataset, named UofTPed50, for benchmarking 3D object detection and tracking of pedestrians. Our focus is on providing an accurate position and velocity benchmark for a pedestrian in the form of GPS ground truth. Currently, a comprehensive benchmark of this type is not publicly available. UofTPed50 consists of 50 sequences of varying distance, trajectory shape, pedestrian appearance, and sensor vehicle velocities. Each sequence contains one pedestrian. The scenarios are broken into five groups: 


\begin{enumerate}
    \item 34 sequences tracking straight lateral trajectories with respect to the stationary car at seven distances performed by three pedestrians. (Sequences 1-34)
    \item 3 sequences tracking straight lateral trajectories with respect to a dynamic car performed by two pedestrians (Sequences 35-37)
    \item 4 sequences tracking straight longitudinal trajectories with respect to the stationary car performed by two pedestrians (Sequences 38-41)
    \item 4 sequences tracking straight longitudinal trajectories with respect to a dynamic car performed by two pedestrians (Sequences 42-45)
    \item 6 sequences tracking complex trajectories (i.e., curves and Zig-Zags) with respect to the stationary car performed by two pedestrians (Sequences 46-51)
\end{enumerate}
Training Sequences 28-37 can be used for tuning the algorithm. They feature one trajectory at each lateral distance. 
\begin{figure} [b]
    \includegraphics[width=\columnwidth]{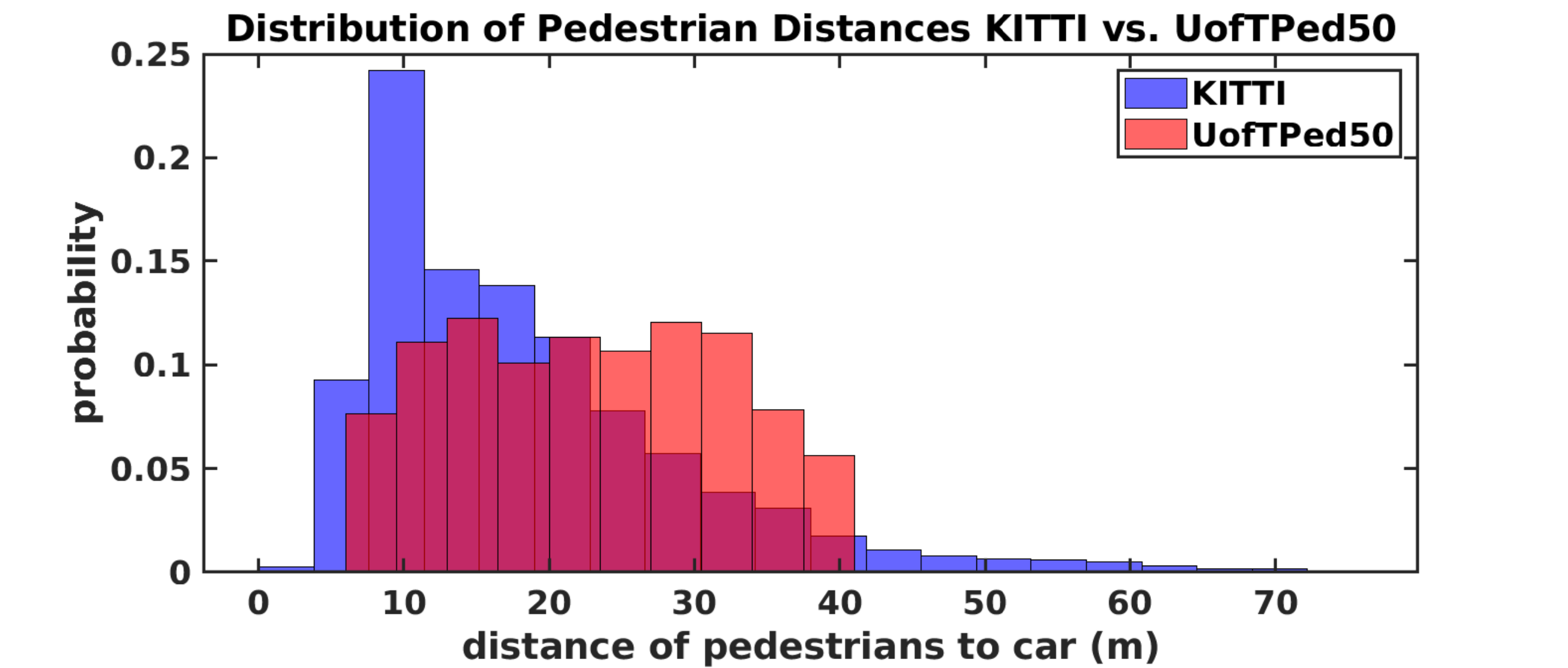}    
    \centering
    \caption{Distribution of pedestrian distances in the KITTI dataset and in the UofTPed50 dataset.}
    \label{fig:distdist}
\end{figure}

We collected UofTPed50 data on our self-driving car, Zeus, illustrated in Figure~\ref{fig:sensors_diagram}. Zeus is a 2017 Chevrolet Bolt Electric Vehicle. Sensor data was collected from a Velodyne HDL-64 LIDAR, a 5 MP Blackfly BFS-U3-51S5C-C monocular camera, and a NovAtel PwrPak7 GPS/IMU with TerraStar corrections (< 10 cm reported position error). Position data for the pedestrian was collected by attaching the tethered antenna of a NovAtel PwrPak7 GPS, also with TerraStar corrections, to the pedestrian's head as illustrated in Figure~\ref{fig:rqt_output}. Ground truth velocity for the pedestrian was generated by smoothing the finite difference between GPS waypoints. To  synchronize data between the car and the pedestrian, we use UTC timestamps. A minor issue observed during the data collection process is that the two GPS units have a constant offset. To correct this, we estimate the centroid of the pedestrian at the beginning of each sequence using LIDAR, and use it to estimate the offset between the two GPS frames. 

There are some key differences between UofTPed50 and the KITTI dataset. First, on KITTI object tracking can only be benchmarked on the 2D image plane. Although there are 3D labels for pedestrians, the multi-object tracking benchmark does not include this information. In UofTPed50, we collect the 3D global position of the ground truth from GPS, which is more consistent in its error. Second, as illustrated in Figure~\ref{fig:distdist}, the KITTI dataset has a narrow distribution of pedestrian distances with nearly 25\% of all labelled pedestrians being roughly 10 m from the car. Despite having a smaller range of distances, UofTPed50 has a more uniform distribution. Finally, most pedestrian sequences in the KITTI dataset contain pedestrians tracking a constant heading. In UofTPed50, we collect sequences with the pedestrian tracking complex curved and Zig-Zag trajectories.

\section{Experiments:} \label{sec:experiments}
In this section, we describe the experimental evaluation we have performed on UofTPed50, as well as on the KITTI Object Tracking benchmark. We used training sequences 28-34 of UofTPed50 to tune our tracker and did not include them in our evaluation. We have divided our evaluation into several subsections. First, we demonstrate the performance of our approach with varying target depth. Second, we show the impact relative motion can have on the accuracy of our position and velocity estimates. Third, we compare the performance of our approach on different trajectory shapes. We also discuss the performance of our approach using several qualitative examples that compare the pedestrian trajectory generated by aUToTrack against the ground truth GPS-based trajectories. We present a qualitative analysis of our velocity tracking performance on several scenarios. We describe the accuracy with which we can assess whether or not a pedestrian is static. Finally, we briefly analyze our performance on the KITTI Object Tracking benchmark.

\subsection{Varying Distance}
The first set of sequences in UofTPed50, the pedestrian walks laterally from one side of the vehicle to the other at evenly spaced distances. Figure~\ref{fig:straight} illustrates an example of a lateral trajectory. We use this to simulate a pedestrian crossing the road immediately in front of the vehicle. We need to ensure that position and velocity estimates remain accurate with increasing distance. Being able to accurately estimate the state of other traffic participants regardless of distance is key to the safety of an autonomous vehicle.

We use Root Mean Squared Error (RMSE) as our error metric for both position and velocity estimation. As summarized in Table~\ref{fig:distdist}, our position and velocity estimation error tends to increase with distance. The position error increases from 0.14 m at 5 m to 0.37 m at 35 m. This appears to be an acceptable increase in error given the range increase. Although we are able to achieve high velocity estimation accuracy up to 30 m, there appears to be a steep drop off in performance at 35 m. This is potentially due to the increasing sparsity of points further from the LIDAR. We have also observed that 30 m is close to the detection range of SqueezeDet trained on KITTI, another potential cause of the drop in performance. 


\begin{table}[H]
    \centering
    \caption{Position and Velocity Estimation Error vs. Target Distance (Sequences 1-27)}
    \label{distance-table}
    \begin{tabular}{|l||l|l|l|l|l|l|l|}
    \hline
    \textbf{\begin{tabular}[c]{@{}l@{}}Target \\ Distance(m)\end{tabular}} & \textbf{5}    & \textbf{10}   & \textbf{15}   & \textbf{20}   & \textbf{25}   & \textbf{30}   & \textbf{35}    \\ \hline \hline
    \textbf{\begin{tabular}[c]{@{}l@{}}Position \\ RMSE (m)\end{tabular}}   & 0.14 & 0.18 & 0.21 & 0.26 & 0.22 & 0.27 & 0.37 \\ \hline
    \textbf{\begin{tabular}[c]{@{}l@{}}Velocity \\ RMSE (m)\end{tabular}} & 0.20 & 0.19 & 0.18 & 0.23 & 0.32 & 0.29 & 0.55 \\ \hline
    \end{tabular}
\end{table}


\subsection{Varying Relative Motion}
In these experiments, the pedestrian is either walking straight towards or away from the vehicle. We repeat the pedestrian trajectories with the vehicle stationary and driving forward. These trajectories are used to simulate a pedestrian walking along a sidewalk. It is important to distinguish between pedestrians moving laterally across and longitudinally along the road. Table~\ref{velocity-table} summarizes the results of this experiment. We note that when the vehicle is driving forward (V > 0) and the pedestrian is walking towards the vehicle, position error increases. In general, high sensor vehicle velocity introduces several complications, such as pointcloud distortion and sensor message misalignment. Thus, our expectation is that error should increase with increasing relative motion. However, it is surprising to see that velocity estimation error remains relatively constant in both cases regardless of vehicle motion. For scenarios where the pedestrian is walking away, relative velocity decreases. As such, we observe that both the position and velocity error also decrease.


\begin{table}[H]
    \centering
    \caption{Position and Velocity Estimation Error vs. Relative Motion (Sequences 38-45)}
    \label{velocity-table}
    \begin{tabular}{|l||l|l|l|l|}
    \hline
    \textbf{Pedestrian Motion}   & \multicolumn{2}{l|}{Walk Towards} & \multicolumn{2}{l|}{Walk Away} \\ \hline \hline
    \textbf{Vehicle Motion}      & V = 0      & V \textgreater~0     & V = 0    & V \textgreater~0    \\ \hline
    \textbf{Position RMSE (m)}   & 0.27       & 0.51                 & 0.31     & 0.27                \\ \hline
    \textbf{Velocity RMSE (m/s)} & 0.21       & 0.19                 & 0.26     & 0.23                \\ \hline
    \end{tabular}
\end{table}

\subsection{Varying Trajectory Shape}
In these experiments, our goal is to push our system in order to find corner cases where performance declines. In the other sequences, the pedestrian motion follows a relatively constant heading. Here, we demonstrate trajectories that have more changes in velocity and direction
Curved trajectories involve the pedestrian jogging in a shape similar to a parabola. Figure~\ref{fig:zig-zag} illustrates an example of a Zig-Zag trajectory. 

Wait trajectories involve the vehicle driving forwards, stopping and then waiting for the pedestrian to cross. Table~\ref{trajectory-table} summarizes the results of this experiment. As expected, the straight-line trajectories (Across, Toward, Away) tend to have the lowest tracking error due to their simplicity. Interestingly, position error remains relatively low during the Curve, Zig-Zag, and Wait trajectories. This can be due to the fact that relative motion remains low. Velocity estimation error increases for the more complex trajectories, but is highest for Zig-Zag. We anticipate this is caused by lag in our current estimator. Estimator parameters were tuned to compromise between lag and smoothness.

\begin{table}[H]
    \centering
    \caption{Position and Velocity Estimation Error vs. Scenario Type (Sequences 46-51)}
    \label{trajectory-table}
    \begin{tabular}{|l||l|l|l|l|l|l|}
    \hline
    \setlength\tabcolsep{4.5pt}
    \textbf{Scenario}                                                       & Curve & Zig-Zag & Across & Toward & Away & Wait \\ \hline \hline
    \textbf{\begin{tabular}[c]{@{}l@{}}Position \\ RMSE(m)\end{tabular}}   & 0.32    & 0.26    & 0.21   & 0.27    & 0.31 & 0.35 \\ \hline
    \textbf{\begin{tabular}[c]{@{}l@{}}Velocity \\ RMSE(m/s)\end{tabular}} & 0.51    & 0.59    & 0.18   & 0.21    & 0.26 & 0.46 \\ \hline
    \end{tabular}
\end{table}

\subsection{Qualitative Analysis}
\begin{figure} [b]
    \includegraphics[width=\columnwidth]{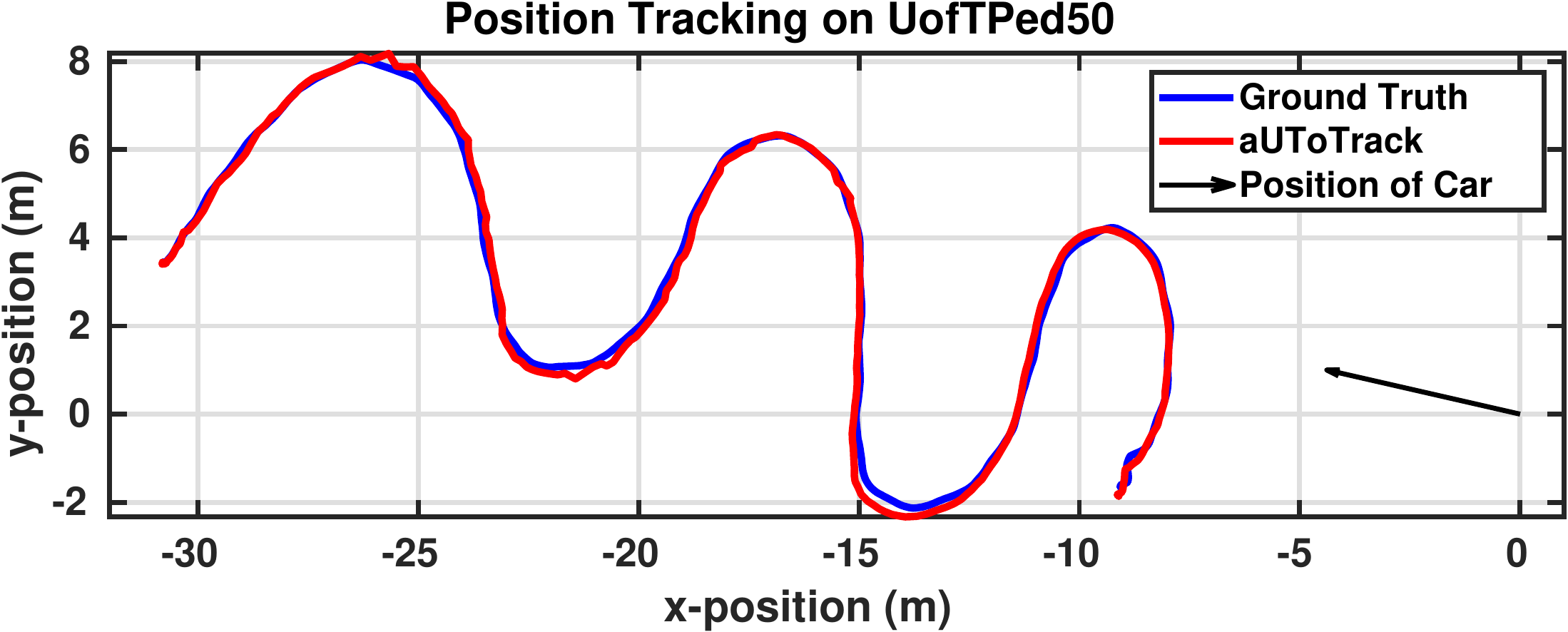}   
    \centering
    \caption{Position tracking for Sequence 51. The pedestrian starts roughly 30m away, then follows a Zig-Zag trajectory towards the stationary car.}
    \label{fig:zig-zag}
\end{figure}

\begin{figure} [t]
    \includegraphics[width=\columnwidth]{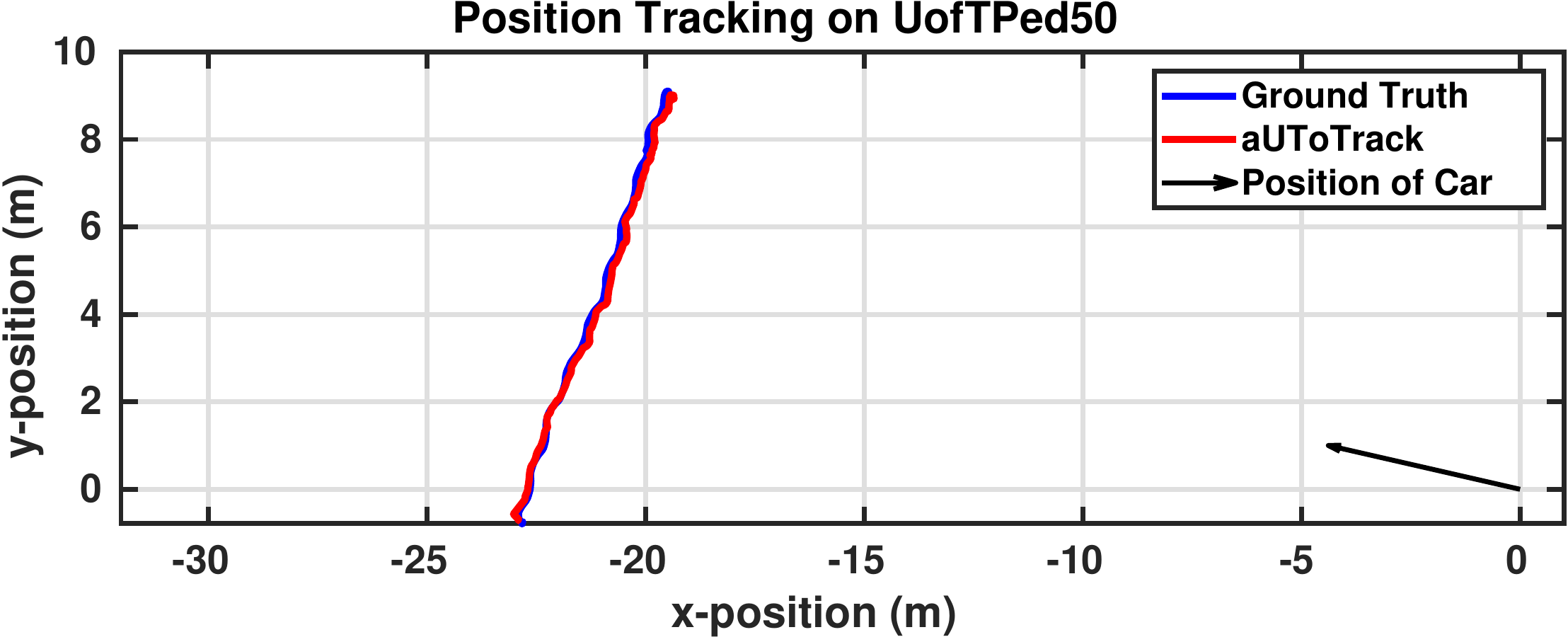}   
    \centering
    \caption{Position Tracking for Sequence 21. The pedestrian starts roughly 20m away on the left of the car, then follows a straight trajectory laterally.}
    \label{fig:straight}
\end{figure}

\begin{figure} [t]
    \includegraphics[width=\columnwidth]{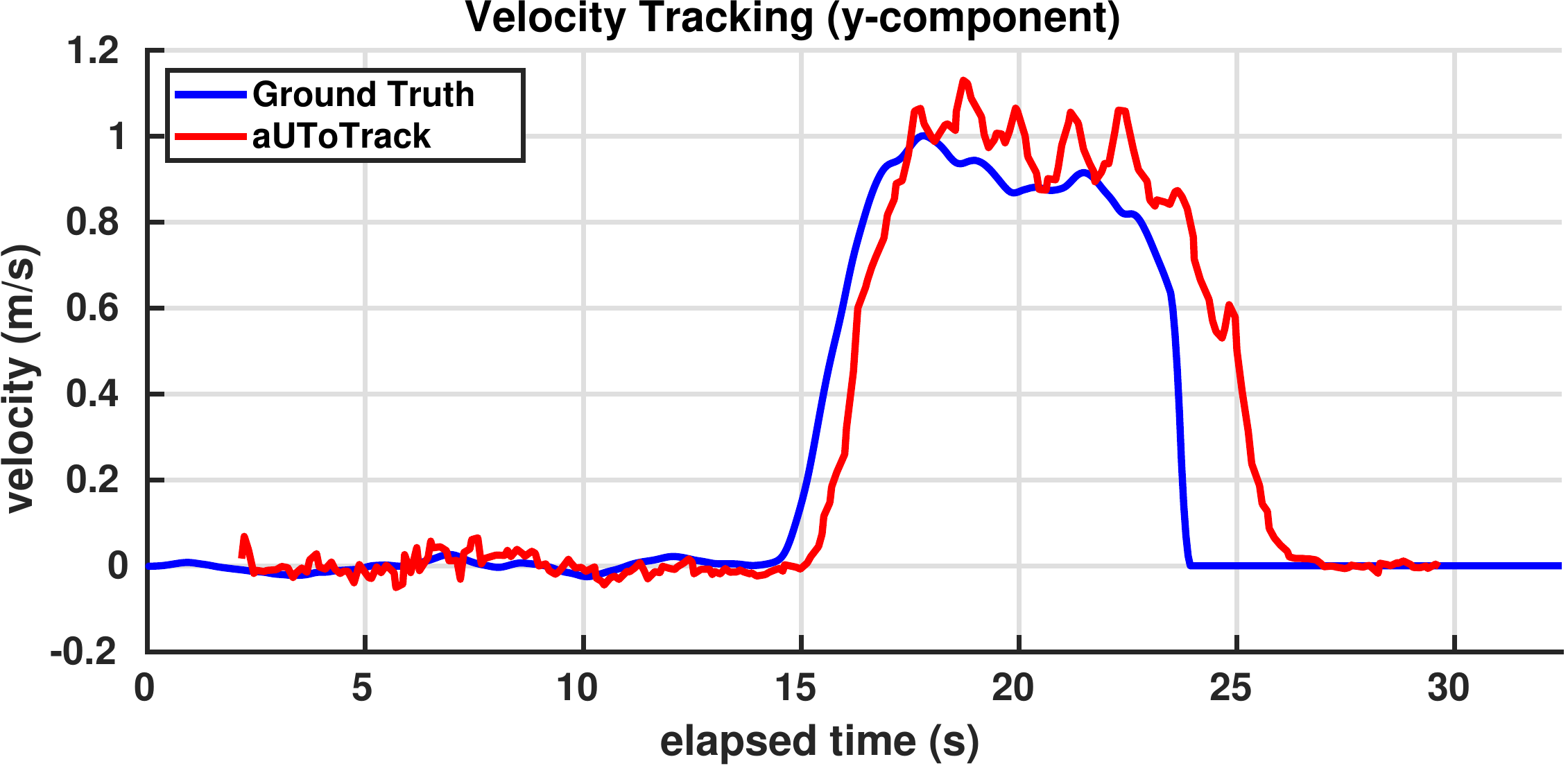}   
    \centering
    \caption{Velocity estimation for Sequence 37. For the first 15 seconds, the car approaches a waiting pedestrian then stops. The pedestrian then crosses laterally and stops.}
    \label{fig:velo-straight}
    \vspace{-5mm}
\end{figure}

\begin{figure} [t]
    \includegraphics[width=\columnwidth]{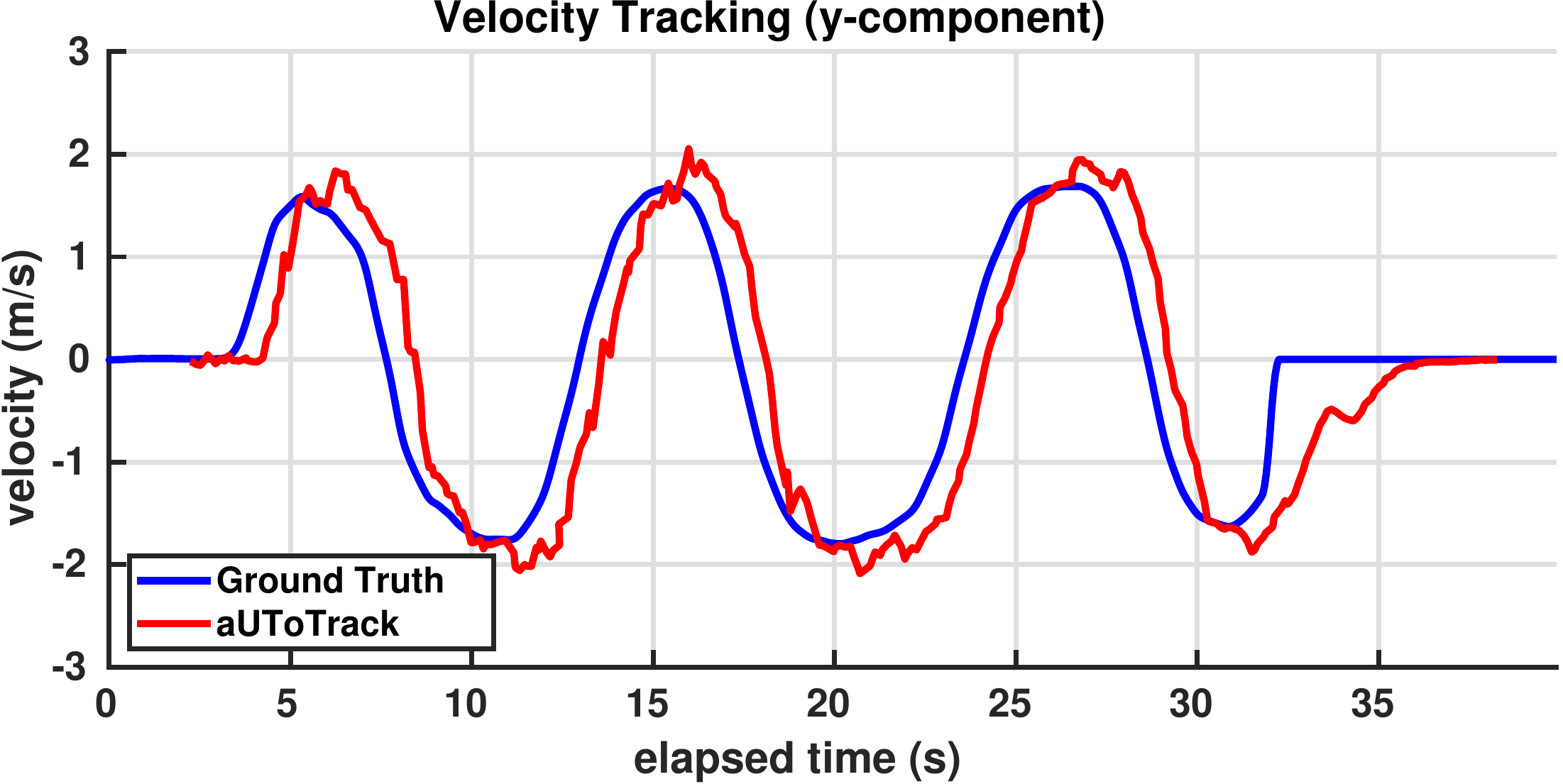}
    \centering
    \caption{Velocity estimation for Sequence 51, corresponding to Figure~\ref{fig:zig-zag}. The pedestrian starts roughly 30m away, then follows a Zig-Zag trajectory towards the stationary car.}
    \label{fig:velo-zig-zag}
    \vspace{-5mm}
\end{figure}

Figures~\ref{fig:straight},~\ref{fig:zig-zag},~\ref{fig:velo-straight},~\ref{fig:velo-zig-zag} directly compare the trajectories and velocities estimated by aUToTrack against the GPS-based ground truth in our dataset. Our first observation is that our approach is capable of replicating the reference position trajectories with high accuracy. However, one can also observe that the position and velocity estimation appears to overshoot and somewhat lag behind the ground truth. This is likely caused by estimator dynamics, and can be remedied with parameter tuning. Even though the Zig-Zag velocities shown in Figure~\ref{fig:velo-zig-zag} change rapidly, we are still able to track the velocity with respectable accuracy. Figure~\ref{fig:velo-straight} shows a similar story, where the pedestrian abruptly changes velocity, challenging the tracker to keep up.

Overall, we are pleased with the performance of aUToTrack on our dataset, although there is clearly room for improvement. Future work will include improving the velocity tracking in challenging scenarios, and boosting the position accuracy at 35 m. 

\subsection{KITTI Object Tracking}
Table~\ref{test-kitti} summarizes the car tracking performance of aUToTrack the KITTI Object Tracking test set. As of writing, we rank among the top five published works. The metrics used in this benchmark are defined as the Clear MOT metrics from \cite{bernardin2008clearmot}. Our approach is very simple: we use LIDAR data and 2D bounding boxes to estimate the 3D position of objects and then we employ baseline data association and tracking techniques. Despite this simplicity, we are able to achieve quite competitive performance, while tieing for highest framerate among the top five published works. It should be noted that we are only able to achieve state-of-the-art performance when using Recurrent Rolling Convolutions (RRC). Since we are working on a tracking benchmark, it seemed appropriate to use a very good detector.

We note that our MT metric ranks quite highly. Our interpretation of this result is that tracking objects in 3D is simply not as difficult as tracking objects within a 2D image plane. Since our system possesses a 3D estimate of the location and velocity of all objects, we are able to predict the motion of objects more reliably than vision-based approaches. We also note that our FRAG and IDS metrics do not do as well. This is easily attributed to the fact that our data association relies purely on the locations of objects in 3D space. We believe that using image features for data association could dramatically reduce these numbers. However, a more complex data association step would likely also increase runtime.


Table~\ref{kitti2}, compares our results on the training set when using different detectors, with and without our 3D clustering information. The results show that choosing the best detector has by far the largest impact on performance on these datasets. Thus, in order to compete with the other approaches listed in Table~\ref{test-kitti}, a powerful detector is required. Table~\ref{kitti2} also show that our addition of 3D information via clustering has a substantial improvement on the MOTA and Precision, a modest improvement to MT, and negligible impact on the other metrics. 

\vspace{-4mm}

\begin{table}[h]
    \centering
    \caption{Results on the KITTI Object Tracking Test Set}
    \label{test-kitti}
    \setlength\tabcolsep{3.5pt}
    \begin{tabular}{|l|l|l|l|l|l|l|l|} 
    \hline
                      & MOTA$\uparrow$          & MOTP$\uparrow$          & MT$\uparrow$            & ML$\downarrow$           & IDS$\downarrow$          & FRAG$\downarrow$         & FPS$\uparrow$         \\ \hline \hline
    TuSimple \cite{choi2015nomt}              & \textbf{86.6} & 84.0          & 72.5          & 6.8          & 293          & 501          & 1.7          \\ \hline
    MOTB.P. \cite{sharma2018beyondpixels}     & 84.2          & \textbf{85.7} & \textbf{73.2} & \textbf{2.8} & 468          & 944          & 3.3          \\ \hline
    IMMDP \cite{xiang2015learning} & 83.0          & 82.7          & 60.6          & 11.4         & 172          & \textbf{365} & 5.3          \\ \hline
    Ours              & 82.3          & 80.5          & 72.6          & 3.5          & 1025         & 1402         & \textbf{100} \\ \hline
    3D-CNN \cite{scheidegger2018mono}            & 80.4          & 81.3          & 62.8          & 6.2          & \textbf{121} & 613          & \textbf{100} \\ \hline
    \end{tabular}
\end{table}

\vspace{-5mm}

\begin{table}[h]
    \centering
    \caption{Ablation Studies: Network Comparison, Use of Clustering On KITTI Object Tracking Training Set}
    \label{kitti2}
    \setlength\tabcolsep{3.5pt}
    \begin{tabular}{|l||l|l|l|l|l|l|l|}
    \hline
    & MOTA$\uparrow$ & MOTP$\uparrow$ & R$\uparrow$ & P$\uparrow$ & MT$\uparrow$ & ML$\downarrow$ & FPS$\uparrow$ \\ \hline \hline
    \textbf{\begin{tabular}[c]{@{}l@{}}SqueezeDet\\+Clustering\end{tabular}} & 48.8            & 78.2            & 64.9              & 85.6                 & 31.4          & 25.7          & \textbf{50}                                                  \\ \hline
    \textbf{SqueezeDet}                                                  & 46.0            & 78.7            & 65.1              & 83.5                 & 31.0          & 26.1          & \textbf{50}                                                  \\ \hline
    \textbf{\begin{tabular}[c]{@{}l@{}}RRC\\+Clustering\end{tabular}}        & \textbf{84.9}   & 79.7   & 94.3     & \textbf{95.0}        & \textbf{87.2} & \textbf{1.9}  & 0.7                                                          \\ \hline
    \textbf{RRC}                                                         & 80.2            & \textbf{80.0}   & \textbf{94.4}     & 92.6                 & 87.0 & \textbf{1.9}  & 0.7                                                          \\ \hline
    \end{tabular}
\end{table}

\vspace{-5mm}


\section{Conclusion}
In this paper, we introduced the UofTPed50 dataset\footnote{\url{autodrive.utoronto.ca/uoftped50} (releasing June 2019)}, an alternative to KITTI for benchmarking 3D Object Detection and Tracking. The UofTPed50 dataset offers precise ground truth for the position and velocity of a pedestrian in 50 scenarios. We also described our approach to the problem of 3D Object Detection and Tracking -- aUToTrack. We use vision and LIDAR to generate raw detections and use GPS/IMU measurements to track objects in a global metric reference frame. We use an off-the-shelf 2D object detector paired with a simple clustering algorithm to obtain 3D position measurements for each object. We then use simple data association and filtering techniques to obtain competitive tracking performance. We demonstrate state-of-the-art performance on the KITTI Object Tracking public benchmark while showing that our entire pipeline is capable of running in less than 75 ms on CPUs only. 

Although lightweight, there are a couple shortcomings to our approach that we highlight here. First, the centroid estimation we employ is not as accurate for cars because clustered points belong to the face of the object facing the LIDAR. For safe autonomous driving on a crowded roadway, a more accurate centroid is likely required. Second, our approach is that the overall recall is bottle-necked by the 2D detector being employed. By training SqueezeDet on a custom dataset from public road driving, we have been able to achieve 85\% precision and 85\% recall for pedestrians on a hold-out validation set. 






Our future work will include improving our velocity estimation on UofTPed50 and reducing the error in our position estimates at ranges exceeding 35 m with our current camera and lens. 


\addcontentsline{toc}{section}{References}
 \printbibliography
\end{document}